\documentclass{nature}

\usepackage{amssymb}
\usepackage{graphicx}
\usepackage{times}
\usepackage{epsfig}
\usepackage{amsmath}
\usepackage{subfigure}

\usepackage{mathtools}
\usepackage{epstopdf}
\usepackage[font=scriptsize]{caption}
\usepackage{float}

\usepackage{algorithm}
\usepackage{algorithmic}

\newcommand{\argmin}{\arg\!\min}

\title{An Iterative Regression Approach for Face Pose Estimation from RGB Images}

\author{Wenye He}

\begin{document}

\maketitle

\begin{abstract}
This paper presents a iterative optimization method, explicit shape regression, for face pose detection and localization. The regression function is learnt to find out the entire facial shape and minimize the alignment errors. A cascaded learning framework is employed to enhance shape constraint during detection. A combination of a two-level boosted regression, shape indexed features and a correlation-based feature selection method is used to improve the performance. In this paper, we have explain the advantage of ESR for deformable object like face pose estimation and reveal its generic applications of the method. In the experiment, we compare the results with different work and demonstrate the accuracy and robustness in different scenarios.  
\end{abstract}

\section*{Introduction}
Pose estimation is an important problem in computer vision, and has enabled many practical application from face expression \cite{ierf1} to activity tracking \cite{ierf2}.  Researchers design a new algorithm called explicit shape regression (ESR) to find out face alignment from a picture \cite{iref1}. Figure 1 shows how the system uses ESR to learn a shape of a human face image. A simple way to identify a face is to find out facial landmarks like eyes, nose, mouth and chin.
\begin{figure*}[h]
\centerline{\epsfig{figure=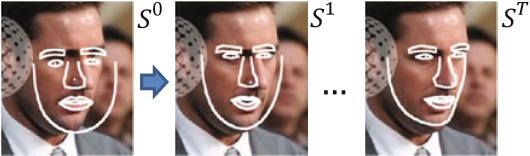,width=10cm}}
\caption{A final shape generated from the inital shape by ESR}
\end{figure*}
The researchers define a face shape $S$ and $S$ is composed of $N_{fp}$ facial landmarks. Therefore, they get $S=[x_1,y_1,...,x_{N_{fp}},y_{N_{fp}}]^T$. The objective of the researchers is to estimate a shape $S$ of a face image. The way to know the accuracy of the estimation is to minimize the alignment error. Equation 1 show how to find alignment error and used for the cascaded learning framework \cite{iref2}. 
\begin{equation}
||S-\hat{S}|_2,
\end{equation}
where $\hat{S}$ is the ground true shape of the image. Nevertheless, the true shape is unknown in testing procedure, so optimization-based method and regression-based method are the most popular approaches for this problem. The researchers use the regression-based method, and, then, they create the algorithm, ESR. Compared to most previous works, ESR does not use any parametric shape models. Considering all facial landmarks are regressed jointly, the researchers train the regressor by reducing the alignment error over training data. This regressor solve large shape variantions and guarantee robustness. Moreover, the researchers design the second regressor to solve small variantion and ensure accuracy. Besides, ESR used the improved version of the cascaded pose regression framework. The layer based regressor has also be adopted for many other applications, such as depth completion \cite{ierf3}. The next section goes over some related works.

\section*{Related Works}
In previous works, active appearance models\cite{iref3} (AAM) is one of influential models in alignment approaches. In 1998, Cootes et al. introduce AAM to interpret an image.  In the same year, they publish how this model interpret face image\cite{iref4} and recognize face\cite{iref5}. In 2007, Saragih and Goechke\cite{iref6} propose a nonlinear discriminative approach to AAM. In 2010, Laurens and Emile\cite{iref7} make an extension of AAM to solve the large variation in face appearance.  In 2011, Sauer and Cootes\cite{iref8} use random forest and boosting regression with AAM. In face alignment, regression-based method is a well-known one. In 2007, Cristinancce and Cootes\cite{iref9} construct constrained local models for face interpretion. In 2010, Valstar et al.\cite{iref10} apply boosted regression find facial features. Cascaded pose regressor (CPR) was originally derived from the boosted regression category \cite{iref2}. CPR has also been successfully used for face detection in videos \cite{ierf7}.

\section*{Method}
This section illustrates the ESR. A basic and essential terms are required to declared here, the normalized shape $M_S\circ S$. $M_S$ is produced from Equation 2 that looks for a $M$ that can make $S$ as close to as the mean shape $\overline{S}$ as possible.
\begin{equation}
M_S=\argmin_M||\overline{S}-M\circ S||_2,
\end{equation}
where $\overline{S}$ is the mean shape and $S$ is an input shape.\\
There are $N$ training samples $\{I_i,\hat{S_i},S^0_i\}^N_{i=1}$, and the stage regressors $(R^1,...,R^T)$. In every Stage $t$, the stage regressor $R^t$ is learnt like Equation 3.
\begin{equation}
\begin{split}
R^t=\argmin_R\sum_{i=1}^N||y_i-R(I_i,S_i^{t-1})||_2\\
y_i=M_{S_i^{t-1}}\circ(\hat{S_i}-S^{t-1}_i),
\end{split}
\end{equation}
where $S^{t-1}_i$ is the estimated shape in previous Stage $t-1$, and $M_{S_i^{t-1}}\circ(\hat{S_i}-S^{t-1}_i)$ is the normalized regression target.\\
In testing, in each Stage $t$, the normalized shape $S^t_i$ is computed as follow,
\begin{equation}
S^t_i=S_i^{t-1}+M^{-1}_{S_i^{t-1}}\circ R^t(I_i,S_i^{t-1}),
\end{equation}
where the normalized shape is updated by the regressor $R^t$ from $S^{t-1}$. The normalization reduces the complication of the regression. Suppose there are two facial images, $I_1$ and $I_2$ with the estimated shape and $I_2$ is transformed from $I_1$. The results of both regressions are different due to transformation. However, normalization, simplifying the problem, would give the same result in both regressions. Algorithem 1 below shows how ESR processing in both training and testing. 
\begin{algorithm}
\begin{spacing}{1.0}
\caption{Explicit Shape Regression (ESR)}
\begin{algorithmic}
\STATE \textbf{Variables:} Training images and labeled shapes $\{I_l,\hat{S_l}\}^L_{l=1}$; ESR model $\{R^t\}^T_{t=1}$; Testing image $I$; predicted shape $S$; $TrainParams$\{times of data augment $N_aug$, number of stages $T$\};
\STATE $TestParams$\{number of multiple initializations $N_{int}$\};
\STATE $InitSet$ which contains exemplar shapes for initialization
\STATE\STATE $\textbf{\textit{ESRTraining}}(\{I_l,\hat{S_l}\}^L_{l=1}, TrainParams, InitSet)$
\STATE // augment training data\\
\STATE $\{I_i, S_i, S_i^0\}^N_{i=1}\gets Initialization(\{I_l,\hat{S_l}\}^L_{l=1}, N_{aug}, InitSet)$
\FOR{$t$ from 1 to $T$}
\STATE $Y \gets \{M_{S_i^{t-1}}\circ(\hat{S_i}-S^{t-1}_i)\}^N_{i=1}$ // compute normalized targets
\STATE $R^t\gets LearnStageRegressor(Y,\{I_l,S^{t-1}_i)\}^N_{i=1})$// using Eq. (3)
\FOR{$i$ from 1 to $N$}
\STATE $S^t_i\gets S^{t-1}_i+M^{-1}_{S_i^{t-1}}\circ R^t (I_l,S^{t-1}_i)$
\ENDFOR
\ENDFOR
\RETURN $\{R^t\}^T_{t=1}$
\STATE\STATE$\textbf{\textit{ESRTesting}}(I,\{R^t\}^T_{t=1}, TestParams, InitSet)$
\STATE //multiple initializations
\STATE $\{I_i,*,S^0_i\}^{N_{int}}_{i=1}\gets Initialization(\{I,*\},N_{int},InitSet)$
\FOR{$t$ from 1 to $T$}
\FOR{$i$ from 1 to $N_{int}$}
\STATE $S^t_i\gets S^{t-1}_i+M^{-1}_{S_i^{t-1}}\circ R^t (I_l,S^{t-1}_i)$
\ENDFOR
\ENDFOR
\STATE $S \gets CombineMultipleResults(\{S^T_i\}^{N_{int}}_{i=1})$
\RETURN $S$
\STATE\STATE$\textbf{\textit{Initialization}}(\{I_c,\hat{S}_c\}^C_{c=1},D, InitSet)$
\STATE $i\gets 1$
\FOR{$c$ from 1 to $C$}
\FOR{$d$ from 1 to $D$}
\STATE $S^0_i\gets$ sampling an exemplar shape from $InitSet$
\STATE $\{I^0_i,\hat{S}^0_i\}\gets\{I_c,\hat{S}_c\}$
\STATE $i\gets i+1$
\ENDFOR
\ENDFOR
\RETURN $\{I^0_i,\hat{S}^0_i,S^0_i\}^{CD}_{i=1}$
\end{algorithmic}
\end{spacing}
\end{algorithm}
In ESR training, the researchers initialize the training data, compute normalized targets and get the stage regressors. In the ESR testing, ther researchers initialize the testing data, normalize targets and output the shape using the regressor.\\
In the initialization, the researchers has an $InitSet$ composed of exemplar shapes which are viewed as representative shapes or groundtruth shapes from the training data. Each exemplar shape generates a number of initial shapes. The researchers' implementation here is 20. The initialization returns a triple $\{I^0_i,\hat{S}^0_i,S^0_i\}^{CD}_{i=1}$, the facial image, the groundtruth shape and the initial shape. Therefore, the number of a triple that has identical facial image and identical groundtruth shape is 20.\\
To improve the performance of the experiments, the researchers introduce two-level boosted regression, external-level and internal-level. The stage regressor $R^t$ is internal-level and is called primitive regressor. The number of iterations is $TK$, where $T$ is the number of iterations in the external-level and $K$ is that in the others. Figure 2 shows the tradeoffs between two level boosted regression. When the total iterations is 5000, the researchers find the lowest mean error as $T=10$ and $K=500$.
\begin{figure*}[h]
\centerline{\epsfig{figure=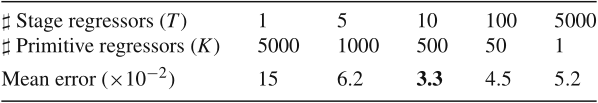,width=12cm}}
\caption{Tradeoffs between two level boosted regression}
\end{figure*}\\
A fern\cite{iref11} is used as the internal-level regressor. Fern can be considered as pixel domain comparison in contrast to features, and can be used for image matching \cite{ierf5}. In the researchers' implementation, a fern contains $F=5$ features. Thus, the feature space and all training samples $\{\hat{y_i}\}^N_{i=1}$ are divided into $2^F$ bins and every bin $b$ means a regression output $y_b$. The prediction of a bin is calculated like Equation 5,
\begin{equation}
y_b=\argmin_y\sum_{i\in\Omega_b}||\hat{y_i}-y||_2,
\end{equation}
where the set $\Omega_b$ implies the samples in the $b$th bin. The best solution is the average,
\begin{equation}
y_b=\frac{\sum_{i\in\Omega_b}\hat{y_i}}{|\Omega_b|}.
\end{equation}
To solve the over-fitting problem, Equation 6 is revised into Equation 7.
\begin{equation}
y_b=\frac{1}{1+\beta/|\Omega_b|}\frac{\sum_{i\in\Omega_b}\hat{y_i}}{|\Omega_b|},
\end{equation}
where $\beta$ is a free shrinkage parameter. Given that the final regressed shape $S$ starts from the initial shape $S^0$ and updates by catching the information from the groundtruth shapes, $S$ is computed like Equation 8,
\begin{equation}
S=S^0+\sum^N_{i=1}w_i\hat{S}_i.
\end{equation}
Before demonstrating the procedure of the internal-boosted regression, shape indexed features is introduced as an important ingredient in the regression. The researchers use pixel-difference features, which means the intensity difference of two pixels in the image. The extracted pixels are unchangeable to similarity transform and normalization. Each pixel is indexed by the local coorinate $\delta^l=(\delta x^l,\delta y^l)$, where $l$ is a landmark associated with the pixel. Considering that pixels indexed by the same global coordinates may be variant because of different face shape while those in the same local coordinates are invariant though in different face shape (detail in Figure 3), 
\begin{figure*}[h]
\centerline{\epsfig{figure=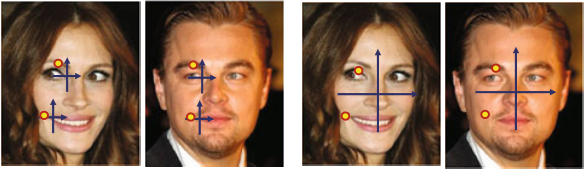,width=12cm}}
\caption{Left pair of images show the local coordinates and right pair of images show the global coordinates. The global coordinates have different meaning due to different images}
\end{figure*}the researchers find the local coordinates first and, then, turn them back to the global coordinates for the intensity difference of two pixels. The transform is showed in Euqation 9.
\begin{equation}
\pi_l\circ S+M^{-1}_S\circ\Delta^l,
\end{equation}
where $\pi_l$ is the operator to get the $x$ and $y$ coordinates of a landmark from the shape. In Figure 3 and the explanation above, different samples has the identical $\delta^l$. Alogorithm 2 shows how the researchers get shape indexed features.\begin{spacing}{1.0}
\begin{algorithm}
\caption{Shape indexed features}
\begin{algorithmic}
\STATE \textbf{Variables:} images and corresponding estimated shapes $\{I_i,S_i\}^N_{i=1}$; number of shape indexed pixel features $P$; number of facial points $N_{fp}$; the range of local coordinate $\kappa$; local coordinates $\{\Delta_\alpha^{l_\alpha}\}^P_{\alpha=1}$; shape indexed pixel features $\rho\in\Re^{N\times P}$; shape indexed pixel-difference features $X\in\Re^{N\times P^2}$;
\STATE\STATE $\textbf{\textit{GenerateShapeIndexedFeatures}}(\{I_i,S_i\}^N_{i=1},N_{fp},P,\kappa)$
\STATE $\{\Delta_\alpha^{l_\alpha}\}^P_{\alpha=1}\gets GenerateLocalCoordinates(FeatureParams)$
\STATE $\rho\gets ExtractShapeIndexedPixels(\{I_i,S_i\}^N_{i=1},\{\Delta_\alpha^{l_\alpha}\}^P_{\alpha=1})$
\STATE $X\gets$ pairwise difference of all columns of $\rho$
\RETURN $\{\Delta_\alpha^{l_\alpha}\}^P_{\alpha=1},\rho,X$

\STATE\STATE$\textbf{\textit{GenerateLocalCoordinates}}(N_{fp},P,\kappa)$
\FOR{$\alpha$ from 1 to $P$}
\STATE $l_\alpha\gets$ randomly drawn a integer in $[1,N_{fp}]$
\STATE $\Delta_\alpha^{l_\alpha}\gets$ randomly drawn two floats in $[-\kappa,kappa]$
\ENDFOR
\RETURN $\{\Delta_\alpha^{l_\alpha}\}^P_{\alpha=1}$

\STATE\STATE$\textbf{\textit{ExtractShapeIndexedPixels}}(\{I_i,S_i\}^N_{i=1},\{\Delta_\alpha^{l_\alpha}\}^P_{\alpha=1})$
\FOR{$i$ from 1 to $N$}
\FOR{$\alpha$ from 1 to $P$}
\STATE $\mu_\alpha\gets\pi_{l_\alpha}\circ S_i+M^{-1}_{S_i}\circ\Delta^{l_\alpha}$
\STATE $\rho_{i\alpha}\gets I_i(\mu_\alpha)$
\ENDFOR
\ENDFOR
\RETURN $\rho$
\end{algorithmic}
\end{algorithm}
\end{spacing}In Algorithm 2, $P$ numbers of pixels are generated, so there are $P^2$ number of pixel-difference features. This makes a huge number of calculations. To optimize the performance and improve the efficiency, the researchers create a correlation-based feature selection to reduce unnecessary computations.\\
The researchers selects $F$ out of $P^2$ features to form a fern regressor. Two properities are required for a good fern: high correlation between each feature in the fern and the regression target and low correlation between features enough to composed complementarily. They propse Equation 10 to maximizing feature's correlation:
\begin{equation}
j_{opt}=argmin_j corr(Y_{prob},X_j),
\end{equation}
where $Y$ is the regression target with $N$ rows and $2N_{fp}$ columns, and $X$ is pixel-difference features matrix with $N$ rows and $P^2$ columns. $N$ is the number of samples. Each column $X_j$ of feature matrix represents a pixel-difference feature. $y_{prob}$ means a projection $Y$ into a column vector from unit Gaussian. The researchers write a pixel-difference feature as $\rho_m-\rho_n$, and they get Equation 10 about the correlation below:
\begin{equation}
\begin{split}
corr(Y_{prob},\rho_m-\rho_n) &= \frac{corr(Y_{prob},\rho_m)-corr(Y_{prob},\rho_n)}{\sqrt{\sigma(Y_{proj})\sigma(\rho_m-\rho_n)}}\\
\sigma(\rho_m-\rho_n) &= cov(\rho_m,\rho_m)+cov(\rho_n,\rho_n)-2cov(\rho_m,\rho_n).
\end{split}
\end{equation}
The pixel-pixel covariances $\sigma(\rho_m-\rho_n)$ can be pre-computed and reused with in each internal-level boosted regression due to fixity of the shape indexed pixels, which reduces the complexity from $O(NP^2)$ to $O(NP)$. Algorithm 3 is the method to selecting correlatioin-based feature.
\begin{spacing}{1.0}
\begin{algorithm}
\caption{Shape indexed features}
\begin{algorithmic}
\STATE \textbf{Input:} regression targets $Y\in\Re^{N\times 2N_{fp}}$; shape indexed pixel features $\rho\in\Re^{N\times N}$; pixel-pixel covariance $cov(\rho)\in\Re^{P\times P}$; number of features of a fern $F$;
\STATE \textbf{Output:} The selected pixel-difference features $\{\rho_{m_f}-\rho_{n_f}\}^F_{f=1}$ and the corresponding indices $\{m_f,n_f\}^F_{f=1}$;
\STATE \STATE $\textbf{\textit{CorrelationBasedFeatureSelection}}(Y,cov(\rho),F)$
\FOR{$f$ from 1 to $F$}
\STATE $v\gets$ randn$(2N_{fp},1)$ // draw a random projection from unit Gaussian
\STATE $Y_{prob}\gets Y_v$ // random projection
\STATE $cov(Y_prob,\rho)\in\Re^{1\times P}\gets$ compute target-pixel covariance
\STATE $\sigma(Y_{prob})\gets$ compute sample variance of $Y_{prob}$
\STATE $m_f=1;n_f=1;$
\FOR{$m$ from 1 to $P$}
\FOR{$n$ from 1 to $P$}
\STATE $corr(Y_{prob},\rho_m-\rho_n)\gets$ compute correlation using Eq.(11)
\IF{$corr(Y_{prob},\rho_m-\rho_n) > corr(Y_{prob},\rho_{m_f}-\rho_{n_f})$}
\STATE $m_f=m;n_f=n;$
\ENDIF
\ENDFOR
\ENDFOR
\ENDFOR
\RETURN $\{\rho_{m_f}-\rho_{n_f}\}^F_{f=1},\{m_f,n_f\}^F_{f=1}$
\end{algorithmic}
\end{algorithm}
\end{spacing}
After introduction of shape indexed features and correlation-based feature selection, the internal-boosted regression is explained in Algorithm 4. The regression consists of $K$ primitive regressors $\{r_1,...,r_K\}$, which are ferns. $F$ thresholds are sampled randomly from an uniform distribution provided the range of pixel difference feature is $[-c,c]$, the range of the uniform distribution is $[-0.2c,0.2c]$. In each iteration a new primitive regressor is learn from the residues left by previous regressors.
\begin{spacing}{1.0}
\begin{algorithm}
\caption{Internal-level boosted regression}
\begin{algorithmic}
\STATE \textbf{Variables:} regression targets $Y\in\Re^{N\times 2N_{fp}}$; training images and corresponding estimated shapes $\{I_i,S_i\}^N_{i=1}$; training parameters $TrainParams\{N_{fp},P,\kappa,F,K\}$; the stage regressor $R$; testing image and corresponding estimated shape $\{I,S\}$;
\STATE\STATE $\textbf{\textit{LearnStageRegressor}}(Y,\{I_i,S_i\}^N_{i=1},TrainParams)$
\STATE $\{\Delta_\alpha^{l_\alpha}\}^P_{\alpha=1}\gets GenerateLocalCoordinates(N_{fp},P,\kappa)$
\STATE $\rho\gets ExtractShapeIndexedPixels(\{I_i,S_i\}^N_{i=1},\{\Delta_\alpha^{l_\alpha}\}^P_{\alpha=1})$
\STATE $cov(\rho)\gets$ pre-compute pixel-pixel covariance
\STATE $Y^0\gets Y$ // initialization
\FOR{$k$ from 1 to $K$}
\STATE $\{\rho_{m_f}-\rho_{n_f}\}^F_{f=1},\{m_f,n_f\}^F_{f=1}\gets \textbf{\textit{CorrelationBasedFeatureSelection}}(Y^{k-1},cov(\rho),F)$
\STATE $\{\theta_f\}^F_{f=1}\gets$ thresholds from an uniform distribution
\STATE $\{\Omega_b\}^{2^F}_{b=1}\gets$ partition training samples into $2^F$ bins
\STATE $\{y_b\}^{2^F}_{b=1}\gets$ randomly drawn two floats in $[-\kappa,kappa]$ compute the outputs of all bins using Eq.(7)
\STATE $r_k\gets\{\{m_f,n_f\}^F_{f=1}\{\theta_f\}^F_{f=1},\{y_b\}^{2^F}_{b=1}\}$ //construct a fern
\STATE $Y_k\gets Y^{k-1}-r^k(\{\rho_{m_f}-\rho_{n_f}\}^F_{f=1})$ //update the targets
\ENDFOR
\STATE $R\gets\{\{r^k\}^K_{k=1},\Delta_\alpha^{l_\alpha}\}^P_{\alpha=1}\}$ //construct stage regressor
\RETURN $R$
\STATE\STATE$\textbf{\textit{ApplyStageRegressor}}(I,S,R)$ // i.e. $R(I,S)$
\STATE $\rho\gets ExtractShapeIndexedPixels(\{I,S\},\{\Delta_\alpha^{l_\alpha}\}^P_{\alpha=1})$
\STATE $\delta S \gets \mathbf{0}$
\FOR{$k$ from 1 to $K$}
\STATE $\delta S\gets \delta S+r^k(\{\rho_{m_f}-\rho_{n_f}\}^F_{f=1})$
\ENDFOR
\RETURN $\delta S$
\end{algorithmic}
\end{algorithm}
\end{spacing}

\section*{Experiments and results}
The researchers compare their approach to previous approaches. Figure 4-6 show samples of images from different datasets.\\\begin{figure*}[h]
\centerline{\epsfig{figure=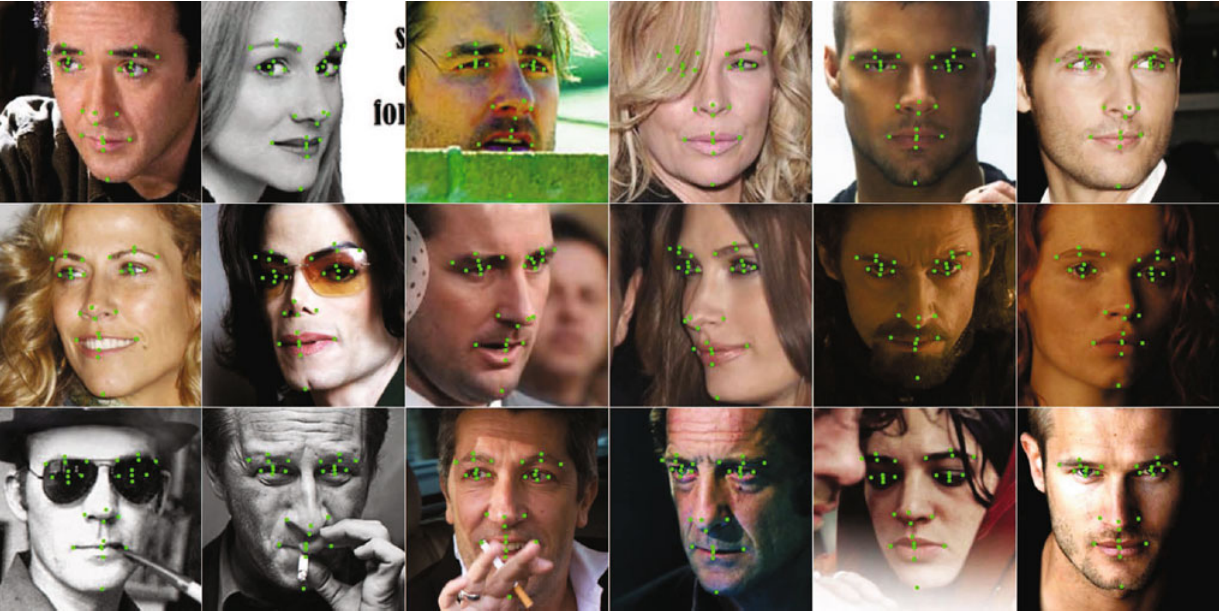,width=10cm}}
\caption{Selected results from LFPW}
\end{figure*}
\begin{figure*}[h]
\centerline{\epsfig{figure=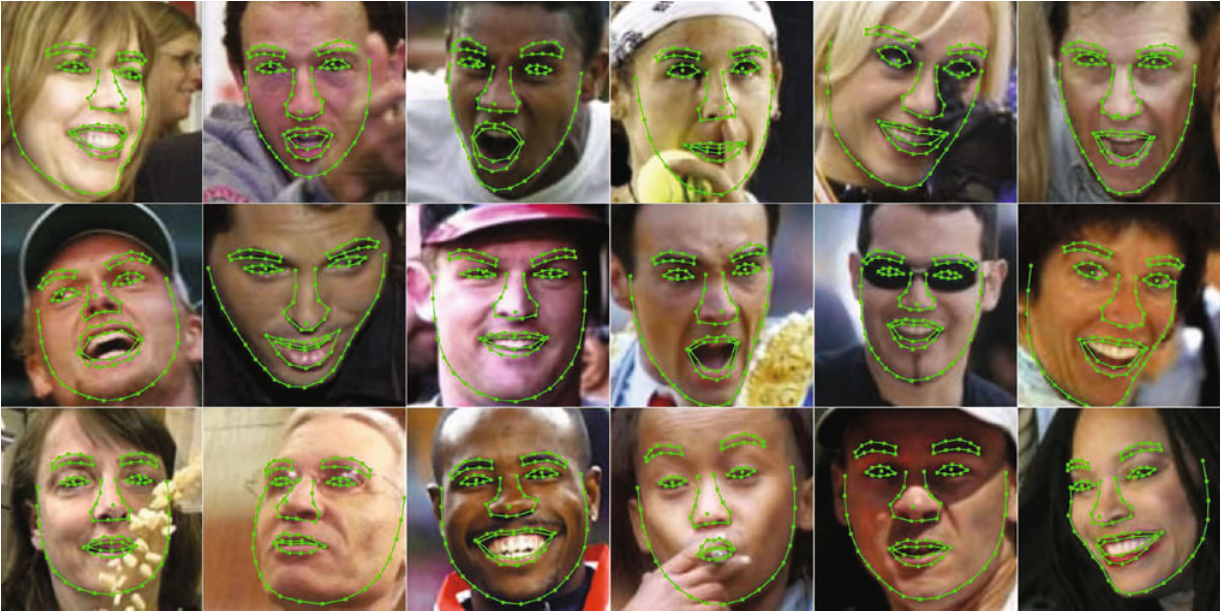,width=10cm}}
\caption{Selected results from LFPW87}
\end{figure*}
\begin{figure*}[h]
\centerline{\epsfig{figure=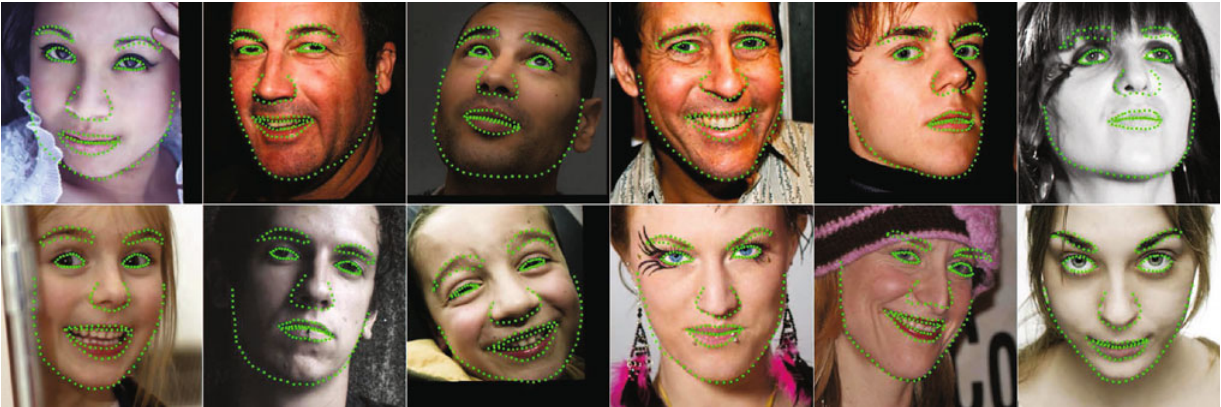,width=10cm}}
\caption{Selected results from Helen dataset}
\end{figure*}
\subsection{Comparison with CE on LFPW}
Compared to the consensus exemplar approach\cite{iref12} on LFPW\cite{iref12}, ESR has more than $10\%$ accurate on most landmarks estimation and smaller overall error.
\begin{figure*}[h]
\centerline{\epsfig{figure=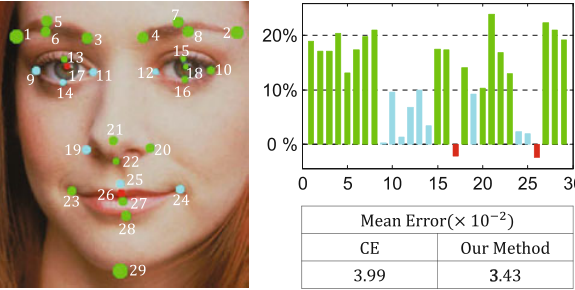,width=10cm}}
\caption{29 facial landmarks on the left image. The radius of circle implies the average error of ESR. The color of the circles suggests accuracy improvement over the CE method. Green has more than $10\%$ accuracy, while cyan has lass than $10\%$ accuracy. The bar graph shows the different accuracy between two methods among all landmarks. The small table shows average error of all landmarks in both methods}
\end{figure*}
\subsection{Comparison with CDS on LFPW87}
A component-based discriminative search (CDS)\cite{iref13} method is proposed by Liang et al.\cite{iref13} ESR beats CDS due to lower error rate.
\begin{figure*}[h]
\centerline{\epsfig{figure=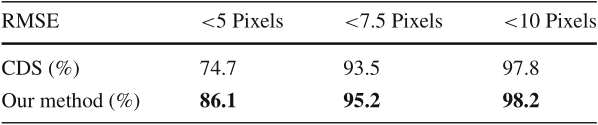,width=10cm}}
\caption{Percentages of test images with root mean square error less than given thresholds on the LFW87 dataset}
\end{figure*}
\subsection{Comparison with STASM and CompASM on Helen}
ESR has around $40\%-50\%$ lower mean error on Helen dataset\cite{iref14} than STASM\cite{iref15} and CompASM\cite{iref14} have.
\begin{figure*}[h]
\centerline{\epsfig{figure=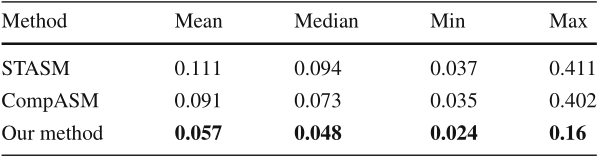,width=10cm}}
\caption{Different errors in ESR, STASM and CompASM}
\end{figure*}

\end{document}